\documentclass{article}

\usepackage[nonatbib,preprint]{neurips_2025}

\usepackage[utf8]{inputenc} 
\usepackage[T1]{fontenc}    
\usepackage{hyperref}       
\usepackage{url}            
\usepackage{booktabs}       
\usepackage{amsfonts}       
\usepackage{nicefrac}       
\usepackage{microtype}      
\usepackage{xcolor}         
\usepackage{multirow}
\usepackage{amsmath}
\usepackage{bm}
\usepackage{graphicx}
\AtBeginDocument{%
  \setlength{\abovedisplayskip}{0pt}
  \setlength{\belowdisplayskip}{0pt}
}

\title{Context-Aware Autoregressive Models for Multi-Conditional Image Generation}

\author{
\hspace{-32pt}
\begin{minipage}{1.1\textwidth}
\centering
  Yixiao Chen\textsuperscript{1}, 
  Zhiyuan Ma\textsuperscript{2}\footnotemark[2]\;\,\footnotemark[1], 
  Guoli Jia\textsuperscript{2},
  Che Jiang\textsuperscript{2},
  Jianjun Li\textsuperscript{4},  
  Bowen Zhou\textsuperscript{2,3}\footnotemark[2] \\
  $^1$ \textnormal{Department of Computer Science and Technology, Tsinghua University,} \\
  $^2$ \textnormal{Department of Electronic Engineering, Tsinghua University,} \\
  $^3$  \textnormal{Shanghai Artificial Intelligence Laboratory,} \\
  $^4$  \textnormal{School of Computer Science and Technology, Huazhong University of Science and Technology} \\
  \tt\small {chenyixi22@mails.tsinghua.edu.cn}, {\tt\small \{mzyth,zhoubowen\}@tsinghua.edu.cn}
\end{minipage}
}

\begin{document}

\maketitle

\renewcommand{\thefootnote}{\fnsymbol{footnote}}
\footnotetext[1]{Zhiyuan Ma leads the project.}
\footnotetext[2]{Corresponding authors.}

\begin{figure}[h]
  \centering
  \includegraphics[width=0.9\linewidth]{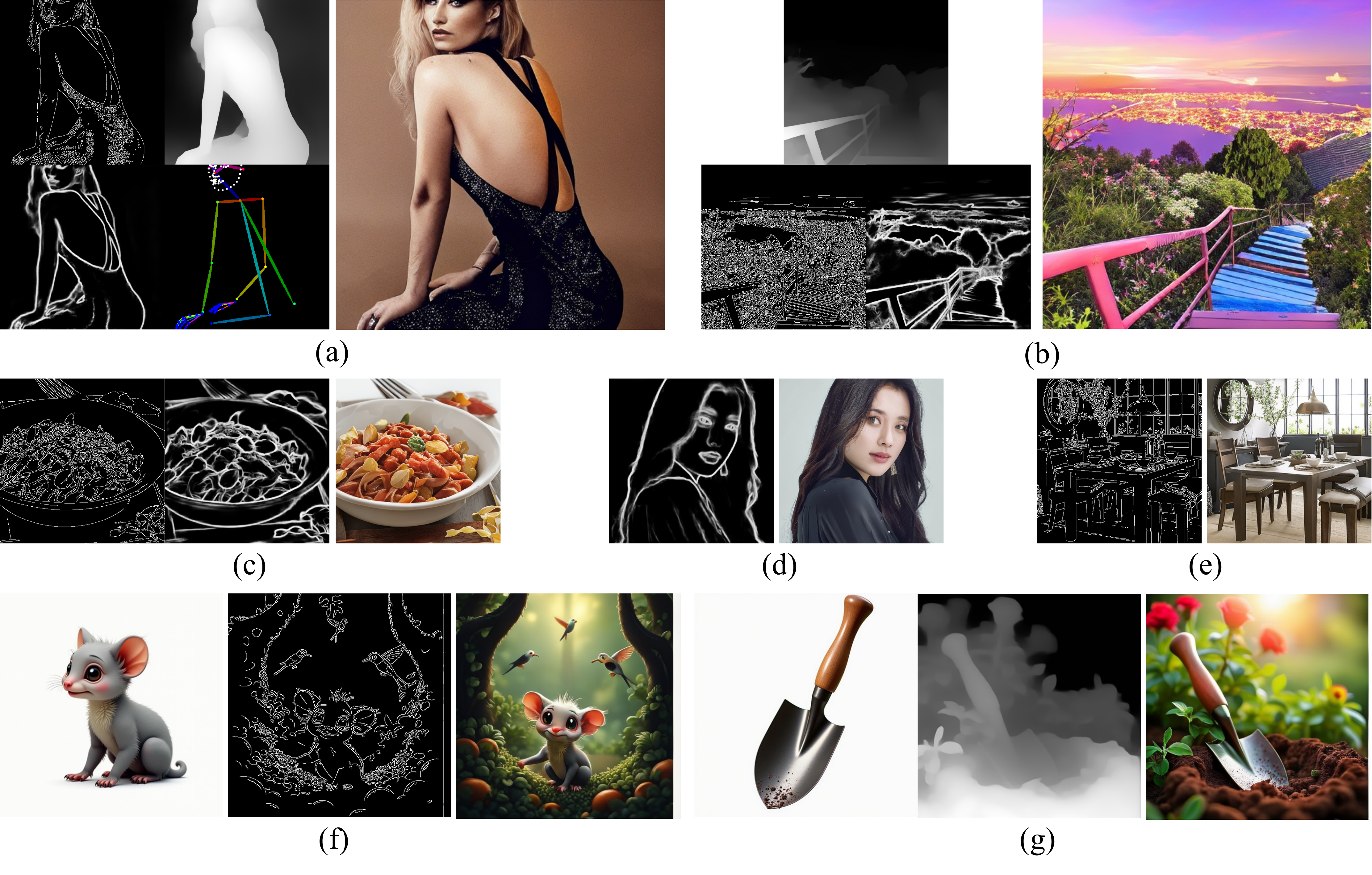}
    \caption{\textbf{Visualization of generated results from our proposed ContextAR framework.} The combinations of conditions are: (a) Canny + Depth + HED + Pose, (b) Canny + Depth + HED, (c) Canny + HED, (d) HED, (e) Canny, (f) Subject + Canny, (g) Subject + Depth. Our framework is implemented on an autoregressive model, achieving excellent controllability while offering remarkable flexibility and versatility.
  }
  \label{fig:demo}
\end{figure}

\begin{abstract}
    Autoregressive transformers have recently shown impressive image generation quality and efficiency on par with state-of-the-art diffusion models. Unlike diffusion architectures, autoregressive models can naturally incorporate arbitrary modalities into a single, unified token sequence—offering a concise solution for multi-conditional image generation tasks. In this work, we propose \textbf{ContextAR}, a flexible and effective framework for multi-conditional image generation. ContextAR embeds diverse conditions (e.g., canny edges, depth maps, poses) directly into the token sequence, preserving modality-specific semantics. To maintain spatial alignment while enhancing discrimination among different condition types, we introduce hybrid positional encodings that fuse Rotary Position Embedding with Learnable Positional Embedding. We design Conditional Context-aware Attention to reduces computational complexity while preserving effective intra-condition perception. Without any fine-tuning, ContextAR supports arbitrary combinations of conditions during inference time. Experimental results demonstrate the powerful controllability and versatility of our approach, and show that the competitive perpormance than diffusion-based multi-conditional control approaches the existing autoregressive baseline across diverse multi-condition driven scenarios. Project page: \href{https://context-ar.github.io/}{https://context-ar.github.io/.}
\end{abstract}

\section{Introduction}

The past few years have witnessed a paradigm shift in visual content generation driven by the advent of generative AI, with diffusion models~\cite{song2020denoising,ho2020denoising,nichol2021improved,dhariwal2021diffusion} rapidly ascending to prominence. These models have launched new state-of-the-art benchmarks in tasks such as text-to-image synthesis~\cite{rombach2022high, flux2024} and conditional image generation~\cite{zhang2023adding, ye2023ip, mou2024t2i}, leveraging iterative noise-refinement procedures to produce high-fidelity images. Despite their remarkable success, diffusion architectures often struggle to integrate heterogeneous conditions—such as textual descriptions, edge maps, and subjects—within a unified framework, compelling researchers to design elaborate, task-specific modifications that compromise both scalability and generality.

In contrast, Transformer-based autoregressive models, celebrated for their in-context learning capabilities in natural language processing (e.g., GPT-style LLMs)~\cite{vaswani2017attention, brown2020language}, naturally accommodate arbitrary sequences of tokens irrespective of modality. By treating text, image features, and auxiliary conditions as a single token stream, these models offer an inherently flexible mechanism for multi-modal reasoning. This observation raises a critical question: \textit{Can autoregressive architectures unlock superior flexibility and generality in multi-conditional image generation, thereby rivaling the adaptability of diffusion-based approaches?}

Existing autoregressive methods for conditional image synthesis remain nascent. Approaches such as EditAR~\cite{mu2025editar} naively prepend condition tokens, resulting in suboptimal control over the generated output. Other methods like ControlAR~\cite{cao2024controllable} circumvent the sequential paradigm entirely by injecting conditions via external adapters, thereby failing to leverage the unified token-based processing. Moreover, these techniques predominantly address single-condition scenarios, leaving the multi-conditional setting largely unexplored.


To bridge this gap, we introduce \textbf{ContextAR}—a \emph{Context-Aware Autoregressive Framework} tailored for multi-conditional image generation. ContextAR stands on the shoulder of the autoregressive paradigm to directly embed diverse conditions  into a unified token sequence for unified representation and generation. Our key contributions are summarized as follows:

\begin{itemize}
    \item \textbf{Unified Sequence Framework:}  We propose a novel framework that integrates multiple conditions as tokens into a single sequence for processing. This design leverages separate embeddings for each type of condition and incorporates hybrid positional encodings, combining Rotary Position Embedding (RoPE)~\cite{su2024roformer} with Learnable Positional Embedding (LPE) to ensure precise spatial alignment and improve condition differentiation.
    
    \item \textbf{Optimized Attention Design}: We propose Cross-Condition Perception Restriction (CCPR) and Intra-Condition Bidirectional Perception (ICBP) to reduce computational complexity while preserving effective intra-condition interactions and causal masking for generation.
    
    \item \textbf{Flexible Condition Composition:} Without any fine-tuning, ContextAR supports arbitrary combinations of conditions during inference time. Empirical results demonstrate that our framework matches or exceeds the controllability of leading diffusion-based methods and outperforms existing autoregressive baseline.

\end{itemize}

\begin{figure}[h]
  \centering
  \includegraphics[width=1.0\linewidth]{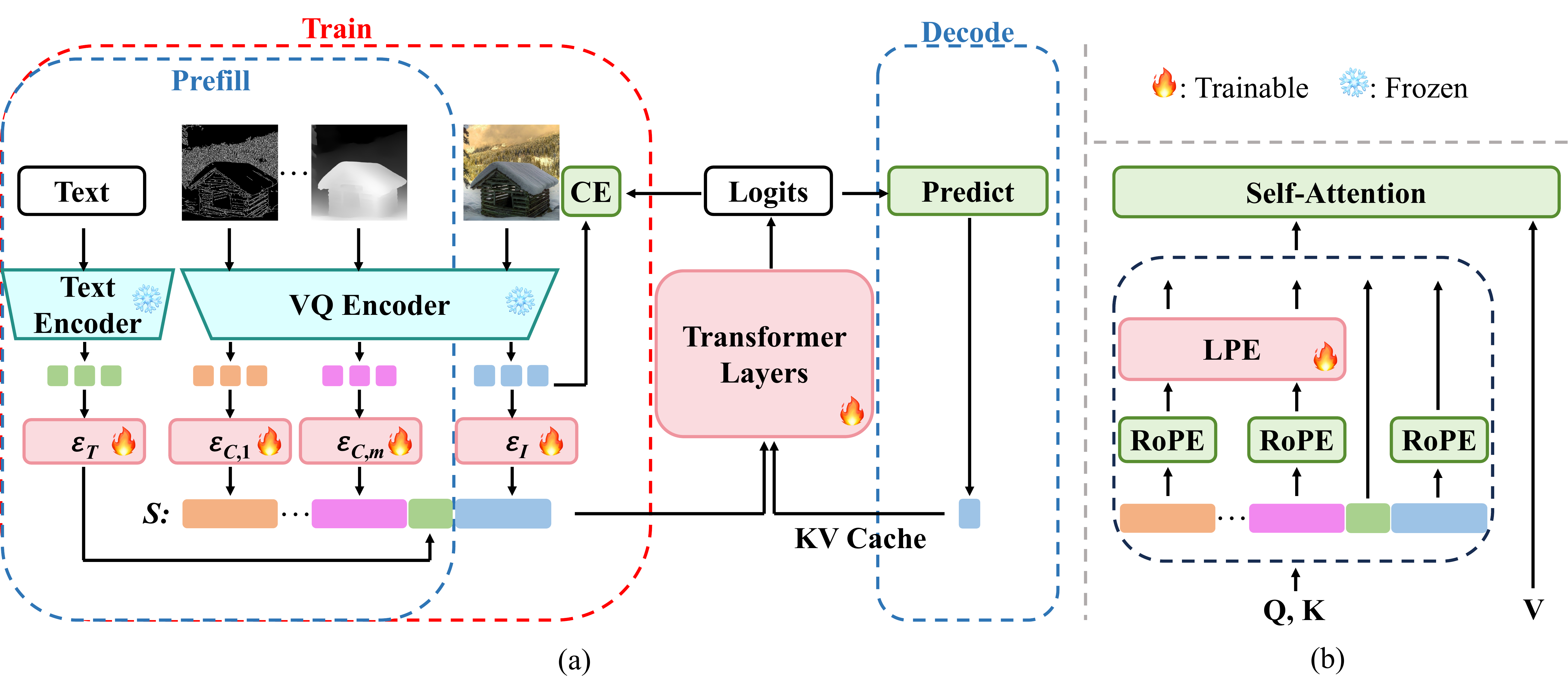}
  \caption{
    Overview of our proposed ContextAR. (a) The overall process of training and inference. Visual conditions, text, and images are incorporated into unified sequence processing. (b) Positional embedding before attention operations.
  }
  \label{fig:overview}
  \vspace{-15pt}
\end{figure}

\section{Related Work}

\subsection{Autoregressive Image Generation}

Autoregressive models~\cite{razavi2019generating,esser2021taming,Parti,Dalle,sun2024autoregressive,li2024autoregressive,tian2024visual,yu2025frequency,wang2025simplear} have gradually achieved performance competitive with diffusion models on image generation tasks. Autoregressive models typically employ a discrete VQ-VAE as a tokenizer~\cite{van2017neural, razavi2019generating,esser2021taming}, encoding images into sequences of tokens through vector quantization, and then generating these sequences using the next-token-prediction paradigm from LLMs. For sequential modeling, raster-scan ordering is most commonly adopted, which arranges two-dimensional tokens into a one-dimensional sequence row by row. Recent works have begun exploring further optimizations of the autoregressive paradigm: VAR~\cite{tian2024visual} employs residual vector quantization, transforming the prediction paradigm into next-scale-prediction to improve image generation quality. MAR~\cite{li2024autoregressive} moves away from conventional vector quantization, instead utilizing continuous tokenizers. FAR~\cite{yu2025frequency} gradually predicts high-frequency details from low-frequency structures, reformulating autoregression as next-frequency-prediction.

In this work, we implement our framework based on LlamaGen~\cite{sun2024autoregressive}, which adopts the Llama2~\cite{touvron2023llama,touvron2023llama2} language model architecture and uses VQ-VAE as its image tokenizer. LlamaGen employs the widely-used raster-scan paradigm and represents a foundational work in autoregressive image generation models.

\subsection{Controllable Text-to-Image Generation}

For image generation, text conditions alone are insufficient to provide the fine-grained control capabilities needed to satisfy user requirements. Based on diffusion models, numerous studies have focused on incorporating visual control signals. For instance, ControlNet~\cite{zhang2023adding} and T2I-Adapter~\cite{mou2024t2i} integrate spatial condition features into diffusion models to generate images with spatial layouts aligned with the conditions. Other approaches such as IP-Adapter~\cite{ye2023ip} and InstantID~\cite{wang2024instantid} introduce subject-type conditions into the generation process, aiming for subject consistency rather than point-to-point spatial alignment. Recently, several diffusion-based multi-conditional control frameworks have been proposed~\cite{qin2023unicontrol,zhao2024uni,hu2023cocktail,wang2025unicombine,pan2025pixelponder}, supporting simultaneous use of multiple conditions to flexibly meet user control requirements.

Meanwhile, condition-controlled generation based on autoregressive models remains at a relatively preliminary stage. EditAR~\cite{mu2025editar} simply appends condition images to the sequence without specialized designs for visual information. ControlAR~\cite{li2024controlar} and CAR~\cite{cao2024controllable} extract features from condition images and subsequently perform fusion. These token-by-token fusion approaches are effective for spatially-aligned tasks but unsuitable for subject-type condition control. Importantly, these methods only support using a single condition at a time.

Our proposed ContextAR flexibly supports multi-conditional control, accommodating both spatially-aligned conditions and subject-driven conditions.

\subsection{Unified Multimodal Autoregressive Models}

Autoregressive models based on transformer architectures possess inherent multimodal compatibility. Recent works~\cite{wu2024janus,wu2024vila,wu2024liquid,team2024chameleon,wang2024emu3,zhou2024transfusion,xie2024show} have focused on unifying text and visual representations as tokens and jointly processing them using transformer architectures for understanding and generation tasks. These approaches demonstrate remarkable flexibility and generality.

\section{Method}
\subsection{Preliminary}

In image generation, autoregressive models often rely on vector quantized variational autoencoders (VQ-VAEs) to transform continuous image data into discrete token sequences. The VQ-VAE encoder maps an input image $\mathcal{I} \in \mathbb{R}^{H \times W \times 3}$ to a latent feature map $\bm{z} \in \mathbb{R}^{h \times w \times d}$, where $h$ and $w$ are reduced spatial dimensions due to downsampling. Each latent vector $\bm{z}_{i,j} \in \mathbb{R}^d$ is then quantized to the nearest entry in a learned codebook $\mathcal{C} = \{\bm{v}_k\}_{k=1}^K$, resulting in a sequence of discrete indices $\bm{q} = (q_1, q_2, \ldots, q_N)$, where $N = h \cdot w$ and each $q_t \in \{1, 2, \ldots, K\}$.

The joint probability of this discrete token sequence, conditioned on some condition $c$, is factorized autoregressively as:

\begin{equation}
p(\bm{q} \mid c) = \prod_{t=1}^{N} p(q_t \mid q_{<t}, c),
\label{eq:ar_image}
\end{equation}
where $q_{<t} = (q_1, q_2, \ldots, q_{t-1})$. In the case of text-to-image generation, the condition $c$ corresponds to the input text prompt.

Training the model involves minimizing the cross-entropy loss of the next-token predictions:
\begin{equation}
\mathcal{L} = -\sum_{t=1}^{N} \log p(q_t \mid q_{<t}, c).
\label{eq:loss}
\end{equation}

LlamaGen~\cite{sun2024autoregressive} adopts a unified sequence formulation by prepending text tokens $\bm{c_T}$ to the image token sequence $\bm{q}$, forming the combined sequence $\bm{S} = [\bm{c_T}, \bm{q}]$. This allows the transformer model to handle multimodal inputs.

During inference, generation proceeds in two stages: \textbf{prefill} and \textbf{decode}. In the prefill stage, all text tokens are processed in parallel, and their attention representations are stored in a KV cache. In the decode stage, the model predicts subsequent tokens step-by-step based on the existing representations in the KV cache.

\subsection{Spatially Overlapped Unified Sequence}

Inspired by the multimodal compatibility of autoregressive models, we integrate visual condition tokens and text condition tokens into the sequence and process this unified sequence with an autoregressive transformer. Formally, we define
\[
\bm{S} = [\bm{c}_1, \bm{c}_2, \ldots, \bm{c}_m,\; \bm{c}_T,\; \bm{q}],
\]
where \(\bm{c}_T\) is the text token sequence, \(\bm{c}_1, \dots, \bm{c}_m\) are the image-condition token sequences, and \(\bm{q}\) is the target image token sequence to be generated. As shown in Figure~\ref{fig:overview}(a), for text condition, we first encode it into features via a pretrained text encoder, then map those features through the model's text embedding layer \(\mathcal{E}_T\) to obtain the text embeddings. For both visual conditions and the target image, we apply a shared pretrained VQ encoder to extract discrete index sequences. Since certain condition types (e.g., Canny edges or HED maps) have token distributions that differ markedly from natural images, we allocate a separate embedding layer \(\mathcal{E}_{C_k}\) for each condition type, initializing its weights from the original image embedding \(\mathcal{E}_I\). After embedding, we obtain a single sequence that simultaneously contains multiple visual conditions, the text prompt, and the target image token embeddings.

\paragraph{Positional Embeddings.}  
In LlamaGen, 2D Rotary Position Embedding (RoPE) is applied to the image tokens. When incorporating condition tokens into the unified sequence, the design of their positional embeddings requires careful consideration.

For text tokens, there is no positional embedding applied, since image generation only requires the semantic content of the text. In contrast, visual condition tokens must carry explicit spatial cues to guide the image synthesis process, so positional embeddings are essential.

A naïve approach is to apply the same 2D RoPE used for image tokens directly to each condition's query and key vectors, causing perfect spatial overlap between condition and image tokens. This ensures that every image token can attend to the condition at the identical spatial location—ideal for alignment-type conditions such as canny or depth maps. However, when multiple conditions are stacked, they all share the same RoPE coordinates, making it difficult for the attention mechanism to distinguish one condition type from another.


To maintain spatial alignment while making positional embeddings distinguishable across different condition types, we apply a hybrid approach. Before each attention computation, we first apply 2D RoPE with the same coordinates as the image to each condition token sequence, ensuring spatial overlap between conditions and images. Subsequently, we apply an additional condition-specific Learnable Positional Embedding (LPE), so that tokens from different condition types receive unique positional offsets. Specifically, for each head \(h\in\{1,\dots,H\}\):
\[
\begin{aligned}
\widetilde{\bm{Q}}_{k}^{(h)} &= \mathrm{RoPE}\bigl(\bm{Q}_{k}^{(h)}\bigr), 
&\quad
\widetilde{\bm{K}}_{k}^{(h)} &= \mathrm{RoPE}\bigl(\bm{K}_{k}^{(h)}\bigr),\\
\widehat{\bm{Q}}_{k}^{(h)} &= \widetilde{\bm{Q}}_{k}^{(h)} + \bm{P}_k, 
&\quad
\widehat{\bm{K}}_{k}^{(h)} &= \widetilde{\bm{K}}_{k}^{(h)} + \bm{P}_k,
\end{aligned}
\]
where:
\(\bm{Q}_{k}^{(h)}\) and \(\bm{K}_{k}^{(h)}\) are the raw query and key projections of condition \(k\) for head \(h\).  
\(\bm{P}_k\in\mathbb{R}^{h\times w \times \frac{d}{H}}\) is the learnable positional offset for condition \(k\), shared identically across every head. Figure~\ref{fig:overview}(b) illustrates the positional embedding process for the entire sequence before the attention operation. After the RoPE+LPE processing, each condition type acquires a unique positional signature, allowing the model to distinguish among multiple stacked conditions during self-attention.  

\begin{figure}[h]
  \centering
  \includegraphics[width=0.9\linewidth]{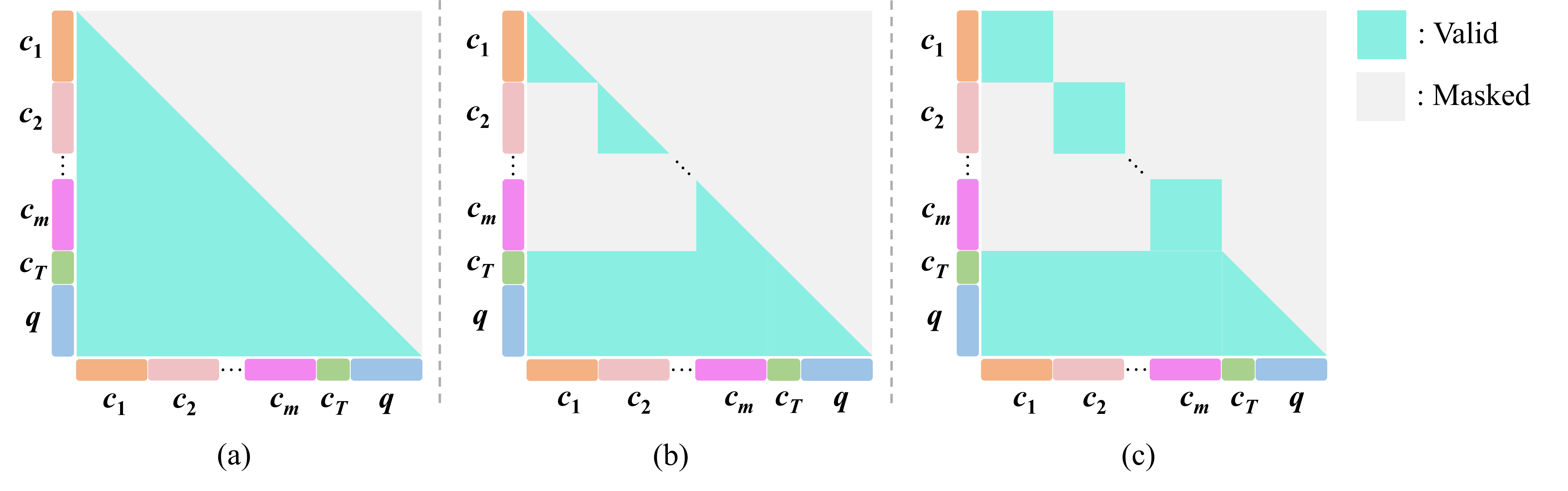}
  \caption{\textbf{Visualization of attention computation.} (a) Normal attention with causal mask, (b) Cross-Condition Perception Restriction, (c) Intra-Condition Bidirectional Perception. 
  }
  \label{fig:attention}
  \vspace{-15pt}
\end{figure}

\subsection{Conditional Context-aware Attention}

\paragraph{Cross-Condition Perception Restriction (CCPR).}
\label{sec:ccpr}
The goal of our approach is to enable the flexible combination of multiple conditions. Once the model is trained on $m$ distinct conditions, it can dynamically select any subset of them to form the conditional input sequence $\bm{S}_{\text{sub}}$ at inference. However, applying naïve self-attention over the unified sequence results in dot-products between queries and keys of different condition types, introducing undesirable coupling and limiting subset selection flexibility.

Additionally, augmenting the sequence $\bm{S}$ with condition tokens incurs high computational cost during both training and prefill phases. Specifically, with $m$ conditions and target sequence $\bm{q}$ of length $N = h \times w$, the self-attention complexity becomes $O\bigl((m+1)^2 N^2\bigr)$ during training and $O\bigl(m^2 N^2\bigr)$ during prefill. The quadratic growth of this cost with the number of conditions makes training with many conditions impractical due to cross-condition attention computations.

To address this, we introduce the Cross-Condition Perception Restriction (CCPR) mechanism, which masks the dot-products between queries and keys of different condition types during self-attention (see Figure~\ref{fig:attention}). This masking prevents inter-condition coupling and reduces the attention complexity from quadratic to linear in terms of the number of conditions, thereby significantly lowering the computational burden.

\paragraph{Intra-Condition Bidirectional Perception (ICBP).}
Autoregressive models typically use causal masks to ensure that each token attends only to its predecessors, maintaining consistency between training and inference. However, for input subsequences processed during the prefill stage—where all tokens are available and computed simultaneously—the strict causality requirement is unnecessary for maintaining this consistency.

Recent studies have shown that bidirectional attention mechanisms outperform causal attention in processing visual information~\cite{xie2024show, zhou2024transfusion, wang2025vision}. Extending the CCPR mechanism, we propose Intra-Condition Bidirectional Perception (ICBP), illustrated in Figure~\ref{fig:attention}(c). ICBP enables bidirectional attention within each condition sequence $\bm{c}_k$ during both training and prefill stages, without the need for masking, thereby improving the model's spatial understanding of conditions. Notably, ICBP preserves causal relationships for image tokens during the decode phase, where sequential token prediction is essential.

\subsection{Training and Inference Strategy}

\paragraph{Training.}  
We freeze both the text encoder and the VQ encoder, and train the model on the full sequence $\bm{S} = [\bm{c}_1, \bm{c}_2, \ldots, \bm{c}_m,\; \bm{c}_T,\; \bm{q}]$. The loss is the cross-entropy of the image tokens:
\[
\mathcal{L} = -\sum_{t=1}^{N} \log p(q_t \mid q_{<t}, c).
\]

To achieve more flexible control, we randomly drop text and condition images during training. For condition $k$, the dropping method is to mask the entire sequence $\bm{c}_k$ in the attention computation. We set the dropout rate to 0.1 for text and 0.25 independently for each visual condition.

\paragraph{Inference.}  
Inference proceeds in two stages. In the prefill stage, feed only the condition sequences and text,  \(\bm{S}' = [\bm{c}_1, \dots, \bm{c}_m, \bm{c}_T]\), into the model and cache all K/V pairs. In the decode stage, autoregressively generate image tokens \(q_t\) one by one.

\paragraph{Classifier-Free Guidance.}  
During inference, we combine unconditional and conditional logits:
\[
l_{\text{CFG}} = l(q_t \mid q_{<t}, \emptyset, \emptyset)
  + \lambda \,\bigl(l(q_t \mid q_{<t}, c_T, c_I)
  - l(q_t \mid q_{<t}, \emptyset, \emptyset)\bigr),
\]
where \(\lambda\) is the guidance scale.

\paragraph{Arbitrary Condition Combinations.}  
The CCPR mechanism (Section~\ref{sec:ccpr}), together with independent masking during training, allows any subset of conditions to be activated at inference time.  
Moreover, our mask-based dropping scheme allows us to omit tokens for unused condition types during inference, avoiding inserting an unconditional embedding and reducing overhead.

\section{Experiments}
\label{sec:experiments}

\subsection{Setup}

\paragraph{Datasets and Baselines.}

We conduct experiments on MultiGen-20M~\cite{qin2023unicontrol} for compositional space-aligned conditions (canny, depth, hed, pose) and compare with several state-of-the-art diffusion-based multi-conditional frameworks~\cite{zhao2024uni,qin2023unicontrol,hu2023cocktail,pan2025pixelponder}. We also compare our method with the autoregressive-based conditional control approach ControlAR~\cite{li2024controlar}, despite its single-condition limitation. To further validate our method, following UniCombine~\cite{wang2025unicombine}, we perform experiments on SubjectSpatial200K combining subject-driven and spatially-aligned conditions (subject-canny, subject-depth).

\paragraph{Metrics.}

We compute FID~\cite{heusel2017gans} and MUSIQ~\cite{ke2021musiq} metrics to evaluate the quality of the generated images. Additionally, we employ SSIM~\cite{wang2004image} to measure the similarity between the generated images and real images, which serves to validate the controllability. For specific condition types, we further utilize dedicated metrics to assess the conditional control capability. In particular, we extract features from the generated images and compute the F1 score for canny, MSE for depth and SSIM for hed, against their respective condition images. For subject conditions, we calculate the CLIP-I~\cite{radford2021learning} score between the generated images and the reference subject images to evaluate their alignment.

\paragraph{Implementation Details.}

We use LlamaGen-XL~\cite{sun2024autoregressive} as our base autoregressive model. All training is done on NVIDIA A100 40GB GPUs. For MultiGen-20M, we use 3 GPUs with a batch size of 1 each and set gradient accumulation to 16 steps, yielding an effective batch size of 48. For SubjectSpatial200K, we use 2 GPUs with a batch size of 2 each and set gradient accumulation to 8 steps, giving a total batch size of 32. We use the AdamW optimizer with a learning rate of 5e-5. Training consists of 30,000 iterations on MultiGen-20M and 25,000 iterations on SubjectSpatial200K. All images are 512×512 pixels. The CFG scale in the inference stage is set to 3.0.

\begin{figure}[h]
  \centering
  \includegraphics[width=0.9\linewidth]{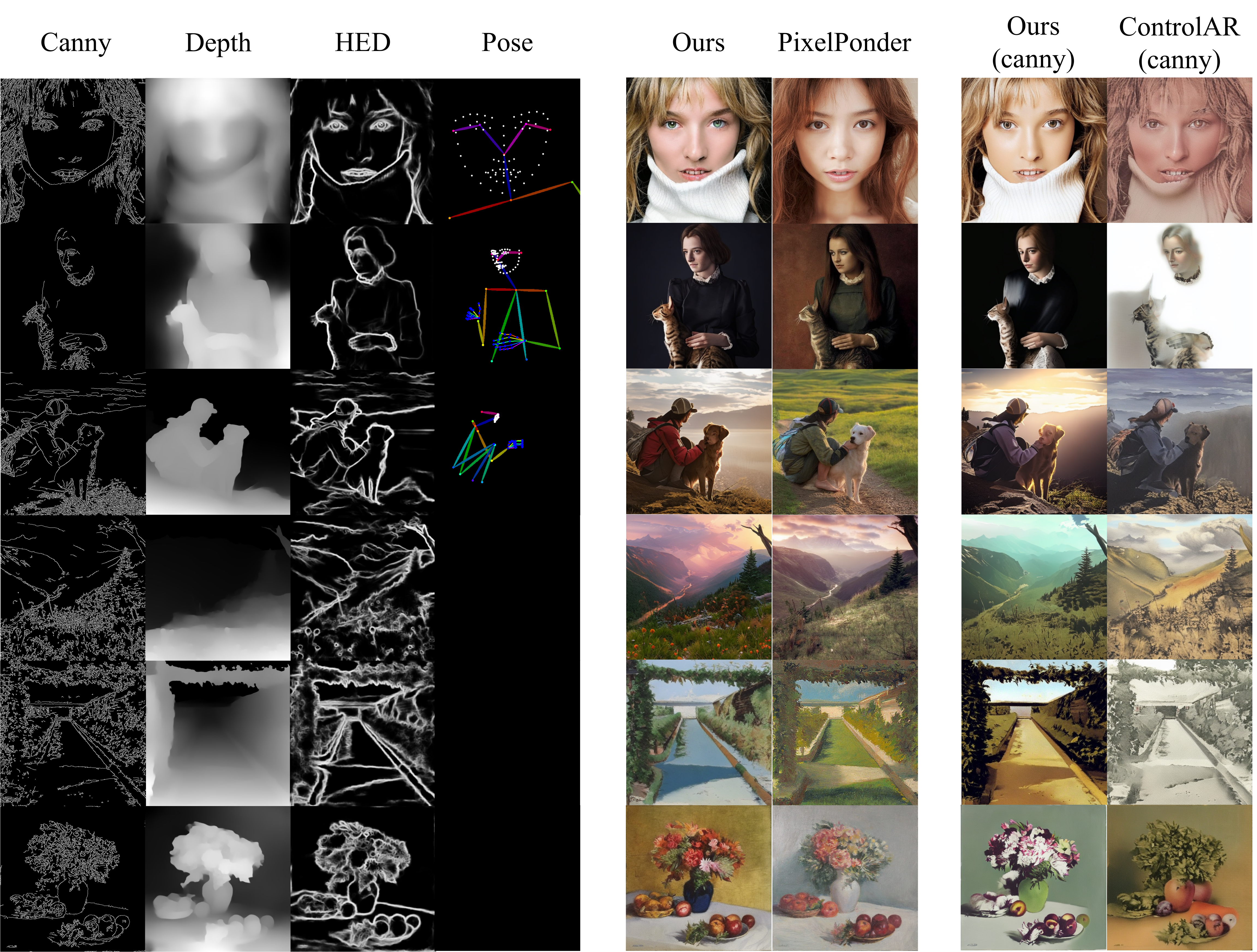}
  \caption{Visualization comparison on MultiGen-20M.}
  \label{fig:main_compare}
  \vspace{-15pt}
\end{figure}

\subsection{Main Results}

\paragraph{Multi-Spatial Conditions}

Table~\ref{tab:multi_spatial_conditions} shows the results of our method on the MultiGen-20M dataset. The SSIM metric indicates that our method demonstrates superior controllability compared to other multi-conditional approaches, with a 21.55\% improvement. Simultaneously, our method achieves the best FID score and the second-best MUSIQ score, proving that it maintains high image quality while providing robust conditional control. Notably, when compared to PixelPonder, which is based on the advanced diffusion model FLUX.1-dev~\cite{flux2024}, our method achieves comparable generation quality despite the underlying autoregressive model having significantly fewer parameters than FLUX.1-dev (775M vs 12B). Figure~\ref{fig:demo}(a) and Figure~\ref{fig:main_compare} visualize the generated results under multiple spatial conditions. Images generated by our method faithfully adhere to various conditional information simultaneously, resulting in high similarity to the referenced original images.

\begin{table}[h]
  \small
  \centering
  \caption{%
    Comparison of generation quality and controllability on the MultiGen-20M dataset. 
    The best results are bolded, and the second-best results are underlined. 
    ``all'' indicates that all conditions (canny, depth, hed, pose) are used.
  }
  \label{tab:multi_spatial_conditions}
  \begin{tabular}{@{}llc cccc@{}}
    \toprule
    \multirow{2}{*}{Methods} & \multirow{2}{*}{Base Model} & \multirow{2}{*}{Conditions} & \multicolumn{3}{c}{Metrics} \\
    \cmidrule(lr){4-6}
    & & & FID $\downarrow$  & SSIM $\uparrow$ & MUSIQ $\uparrow$ \\
    \midrule
    Uni-ControlNet~\cite{zhao2024uni}   & SD1.5  & all  & 32.58  & 29.37  & 65.85  \\
    UniControl~\cite{qin2023unicontrol} & SD1.5  & all  & 25.15  & 35.58  & \textbf{72.05}  \\
    Cocktail~\cite{hu2023cocktail}      &  SD2.1 & pose+hed  & 24.67  & 24.14  & 67.13  \\
    PixelPonder~\cite{pan2025pixelponder} &  FLUX.1-dev  & all  & \underline{11.85}  & \underline{43.99}  & 69.54  \\
    \midrule
    Ours                              & LlamaGen-XL  & all  & \textbf{10.42}  & \textbf{53.47}  & \underline{70.35}  \\
    \bottomrule
  \end{tabular}
  \vspace{-15pt}
\end{table}

\paragraph{Comparision with ControlAR.}

Our method differs from another autoregressive-based control method ControlAR~\cite{li2024controlar}, which is a single-condition method, meaning that only one condition type can be used at a time. In Table~\ref{tab:campare_with_controlar}, we directly compare our model trained in a multi-conditional setting with ControlAR by using only a single condition type as input. As shown, even without specific training for individual conditions, our method still outperforms ControlAR. Furthermore, we demonstrate that combining multiple condition types maintains high control capabilities across condition-specific metrics while achieving consistent FID reductions as more conditions are added. This confirms the flexibility of our approach and the effectiveness of our joint conditional control mechanism. Figure~\ref{fig:demo}(b)(c)(d)(e) and Figure~\ref{fig:main_compare} shows the generation results of our method under various condition combinations.

\begin{table}[h]
  \small
  \centering
  \caption{
    Comparison of generation quality and controllability on the MultiGen-20M dataset. Our method and ControlAR are both based on the same LlamaGen-XL model. Note that our model is jointly trained on four conditions (canny, depth, hed, pose) on MultiGen-20M, and can directly select a subset of conditions during inference without fine-tuning for specific condition combinations.
  }
  \label{tab:campare_with_controlar}
  \begin{tabular}{@{}lc cccc@{}}
    \toprule
    \multirow{2}{*}{Methods} & \multirow{2}{*}{Conditions} & \multicolumn{3}{c}{Metrics} \\
    \cmidrule(lr){3-5}
    & & \multicolumn{1}{c}{FID $\downarrow$}  & \multicolumn{1}{c}{F1 $\uparrow$} & \multicolumn{1}{c}{SSIM(hed) $\uparrow$} \\
    \midrule
    ControlAR~\cite{li2024controlar}   & canny  & 30.44  & 0.31 & -  \\
    Ours                               & canny  & \textbf{18.53}  & \textbf{0.34} & -\\
    \midrule
    ControlAR                          & hed  & 12.62  & -  & 83.48 \\
    Ours                               & hed  & \textbf{11.86}  & -  & \textbf{83.52} \\
    \midrule
    Ours                               & canny+hed  & 11.09   & 0.33  & 83.45  \\
    Ours                               & all  & 10.42   & 0.33  & 83.53  \\
    \bottomrule
  \end{tabular}
  \vspace{-15pt}
\end{table}

\begin{figure}[h]
  \centering
  \includegraphics[width=0.9\linewidth]{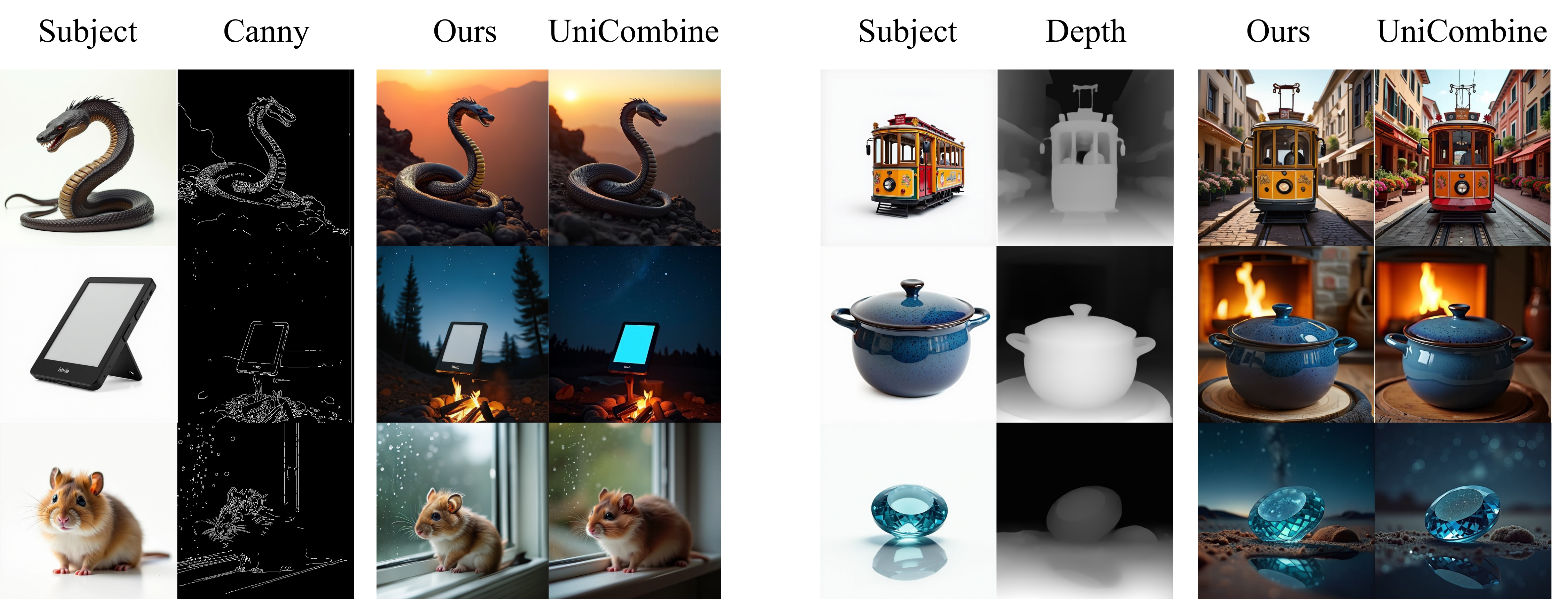}
  \caption{Visualization comparison on SubjectSpatial200K.}
  \label{fig:subject_spatial}
  \vspace{-15pt}
\end{figure}

\paragraph{Subject-Spatial Conditions.}

Subject-driven conditions differ from spatially-aligned conditions in that they do not require precise point-to-point spatial control, but rather semantic alignment. In Table~\ref{tab:campare_with_unicombine}, we compare our method with UniCombine under the combined subject-canny and subject-depth conditions. The F1, MSE, and CLIP-I scores demonstrate that our method can simultaneously maintain high semantic consistency and spatial alignment. Figure~\ref{fig:demo}(f)(g) and Figure~\ref{fig:subject_spatial} showcase our method's generation results for this task. Our approach is effective for both spatial-type conditions and subject-type conditions, demonstrating high generality.

\begin{table}[h]
  \small
  \centering
  \caption{
    Comparison of generation quality and controllability on the SubjectSpatial200K dataset. 
  }
  \label{tab:campare_with_unicombine}
  \begin{tabular}{@{}llc cccc@{}}
    \toprule
    \multirow{2}{*}{Methods} & \multirow{2}{*}{Base Model} & \multirow{2}{*}{Conditions} & \multicolumn{4}{c}{Metrics} \\
    \cmidrule(lr){4-7}
    & & & \multicolumn{1}{c}{FID $\downarrow$} & \multicolumn{1}{c}{F1 $\uparrow$} & \multicolumn{1}{c}{MSE $\downarrow$} & \multicolumn{1}{c}{CLIP-I $\uparrow$} \\
    \midrule

    UniCombine~\cite{wang2025unicombine}  & FLUX.1-schnell    & subject+canny & \textbf{6.01}   & 0.24  & - & 78.92 \\
    Ours          & LlamaGen-XL   & subject+canny  & 6.62  & \textbf{0.30}  & - & \textbf{79.27} \\
    \midrule
    UniCombine & FLUX.1-schnell & subject+depth  & 6.66   & -  & 196.65  & 79.01  \\
    Ours           & LlamaGen-XL   & subject+depth  & \textbf{6.55}   & -  & \textbf{165.56} & \textbf{79.22}  \\
    \bottomrule
  \end{tabular}
  \vspace{-15pt}
\end{table}

\subsection{Ablation Study}

\paragraph{CFG Scale.}

We evaluated the impact of different CFG scale values on controllability and generation quality. As shown in Table \ref{tab:cfg_scale}, the SSIM increases with larger CFG scales, indicating improved conditional control. This validates the effectiveness of our condition drop strategy during training. However, excessively large CFG values lead to degraded generation quality. Therefore, we select CFG = 3.0 to balance controllability and image fidelity.

\setlength{\belowcaptionskip}{2pt}
\begin{minipage}{\textwidth}
\begin{minipage}[t]{0.47\textwidth}
  \makeatletter\def\@captype{table}
  \small
  \centering
  \caption{
    Comparison of differnet CFG scale on the MultiGen-20M dataset. Here all conditions (canny, depth, hed, pose) are used.
  }
  \label{tab:cfg_scale}
  \begin{tabular}{@{}c ccc@{}}
    \toprule
    \multirow{2}{*}{CFG} & \multicolumn{3}{c}{Metrics} \\
    \cmidrule(lr){2-4}
    & \multicolumn{1}{c}{FID $\downarrow$} & \multicolumn{1}{c}{SSIM $\uparrow$} & \multicolumn{1}{c}{MUSIQ $\uparrow$}\\
    \midrule

    1.0 &  9.29 & 50.02 & 69.61 \\
    2.0 &  9.79 & 52.94 & 70.28 \\
    3.0 & 10.42 & 53.47 & 70.35 \\
    4.0 & 10.85 & 53.65 & 70.32 \\
    7.5 & 11.46 & 53.66 & 70.33 \\
    \bottomrule
  \end{tabular}
  
  \label{sample-table}
\end{minipage}
\hfill
\begin{minipage}[t]{0.47\textwidth}
  \makeatletter\def\@captype{table}
  \small
  \centering
  \caption{
    Ablation of positional embedding on the SubjectSpatial200K dataset. Here the conditions used are subject and depth.
  }
  \label{tab:position_embedding}
  \begin{tabular}{@{}l ccc@{}}
    \toprule
    \multirow{2}{*}{Methods} & \multicolumn{3}{c}{Metrics} \\
    \cmidrule(lr){2-4}
    & \multicolumn{1}{c}{FID $\downarrow$} & \multicolumn{1}{c}{MSE $\downarrow$} & \multicolumn{1}{c}{CLIP-I $\uparrow$}\\
    \midrule

    RoPE & 6.67 & 167.09 & 79.11 \\
    RoPE+LPE & 6.55 & 165.56 & 79.22 \\
    \bottomrule
  \end{tabular}
\end{minipage}
\end{minipage}

\paragraph{Positional Embedding.}

To validate the effectiveness of our RoPE+LPE approach for condition images, we compared this strategy with simply applying the same RoPE to both the condition sequence $\bm{c_k}$ and the image sequence $\bm{q}$. The results are shown in Table~\ref{tab:position_embedding}. As observed, the RoPE+LPE strategy achieves more precise control.

\begin{figure}[h]
    \centering
    \begin{minipage}{0.45\textwidth}
      \centering
      \includegraphics[width=0.85\linewidth]{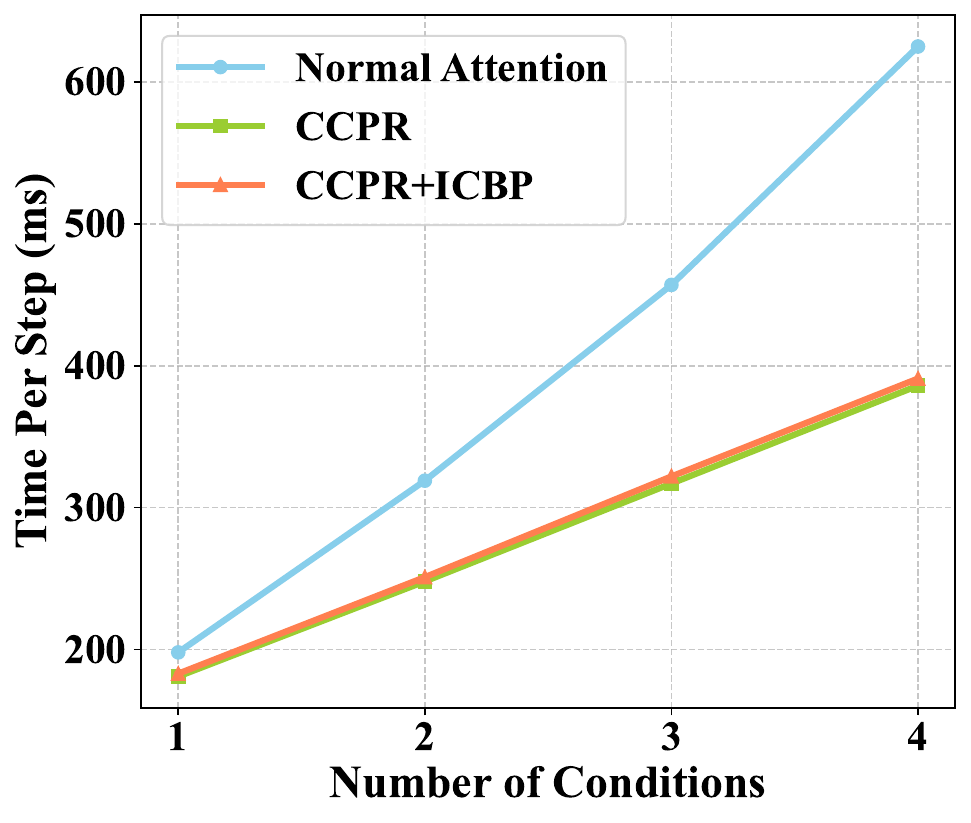}
      \caption{Comparison of different attention mechanisms on the training time cost. Here the batch size is 1.}
      \label{fig:attention_compute}
    \end{minipage}
    \hfill
    \begin{minipage}{0.44\textwidth}
      \centering
      \includegraphics[width=0.85\linewidth]{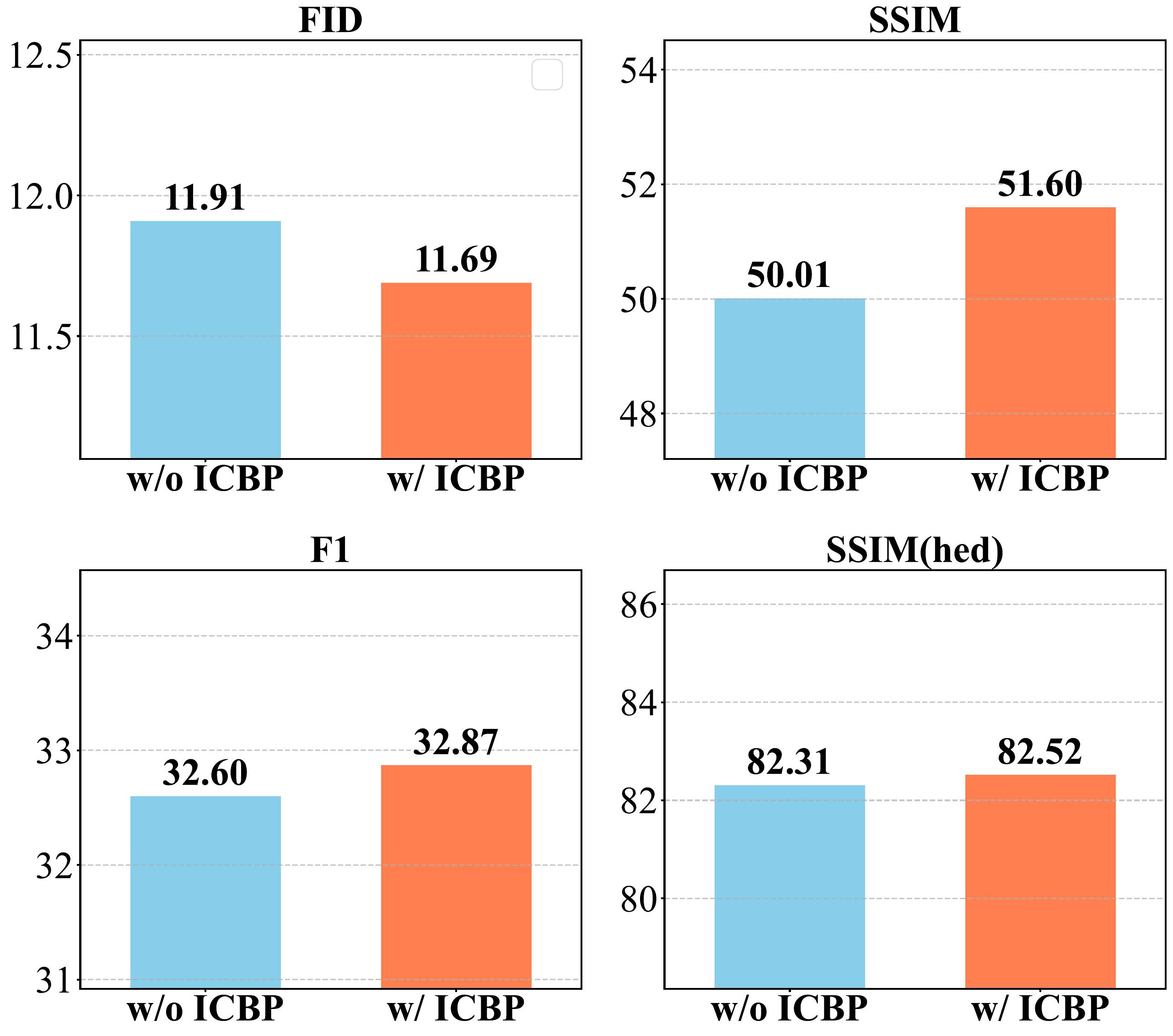}
      \caption{Ablation of ICBP on the MultiGen-20M dataset. Here all conditions are used. Both methods are trained for 10k iterations.}
      \label{tab:ablation_attention}
    \end{minipage}
    \vspace{-15pt}
  \end{figure}

\paragraph{Attention Design.}
  We evaluated three attention mechanisms (standard causal attention, CCPR, and CCPR+ICBP) in terms of per-step computation speed during the training phase. As shown in Figure~\ref{fig:attention_compute}, as the number of conditions increases, the computational cost with the CCPR mechanism is significantly lower than that of standard attention, demonstrating the effectiveness of our approach. Moreover, the bidirectional attention introduced in ICBP adds almost no additional computational overhead.
  Furthermore, we examined the impact of ICBP on conditional control capability. The results in Table~\ref{tab:ablation_attention} show that incorporating the ICBP mechanism improves controllability, confirming that bidirectional attention is beneficial for understanding visual modality conditional information.

\section{Conclusion}

We introduce \textbf{ContextAR}, a flexible autoregressive framework for multi-conditional image generation, which integrates diverse visual conditions directly into a unified token sequence. By combining hybrid positional encodings and Conditional Context-aware Attention, ContextAR maintains spatial alignment, enhances inter-condition discrimination, and significantly reduces computational complexity. Experiments demonstrate its superior controllability, versatility, and competitive performance compared to diffusion-based and autoregressive baselines across various multi-condition scenarios.

\begin{ack}
This work is supported by the National Science and Technology Major Project (2023ZD0121403), National Natural Science Foundation of China (No. 62406161), China Postdoctoral Science Foundation (No. 2023M741950), and the Postdoctoral Fellowship Program of CPSF (No. GZB20230347).
\end{ack}

\bibliographystyle{unsrt}
\bibliography{reference}

\end{document}